%% file: paper.tex
\documentclass[sigconf,screen]{acmart}
\setcopyright{cc}
\setcctype{by}
\acmJournal{PACMMOD}
\acmYear{2026} \acmVolume{4} \acmNumber{1 (SIGMOD)} \acmArticle{54} \acmMonth{2} \acmPrice{}
\acmDOI{10.1145/3786668}
\received{July 2025}
\received[revised]{October 2025}
\received[accepted]{November 2025}

\input{meta/packages}

\input{meta/commands}

\begin{document}

\title{\mixtera: A Data Plane for Foundation Model Training}

\settopmatter{authorsperrow=3}

\author{Maximilian Böther}
\orcid{0000-0003-4093-4361}
\email{mboether@ethz.ch}
\affiliation{%
  \institution{ETH Zurich}
  \country{Switzerland}
}

\author{Xiaozhe Yao}
\orcid{0000-0002-4661-533X}
\email{xiaozhe.yao@ethz.ch}
\affiliation{%
  \institution{ETH Zurich}
  \country{Switzerland}
}

\author{Tolga Kerimoglu}
\orcid{0009-0004-1175-338X}
\email{tkerimoglu@student.ethz.ch}
\affiliation{%
  \institution{ETH Zurich}
  \country{Switzerland}
}

\author{Dan Graur}
\orcid{0009-0001-0682-2422}
\email{dan.graur@ethz.ch}
\affiliation{%
  \institution{ETH Zurich}
  \country{Switzerland}
}

\author{Viktor Gsteiger}
\orcid{0000-0002-6750-5500}
\email{vgsteiger@student.ethz.ch}
\affiliation{%
  \institution{ETH Zurich}
  \country{Switzerland}
}

\author{Ana Klimovic}
\orcid{0000-0001-8559-0529}
\email{aklimovic@ethz.ch}
\affiliation{%
  \institution{ETH Zurich}
  \country{Switzerland}
}

\begin{abstract}
    \input{content/1_abstract}
\end{abstract}

\maketitle

\input{content/2_intro}
\input{content/3_background}

\input{content/4_challenges}

\input{content/5_design}
\input{content/6_impl}

\input{content/7_eval}

\input{content/8_conclusion}

\begin{acks}
We thank Beste Güney for her contributions to \mixtera's codebase.
We thank  Antoni-Joan Solergibert i Llaquet, Imanol Schlag, Steven Hand, Martin Jaggi, Antoine Bosselut, Alexander Hägele, Loubna Ben Allal, Quentin de Laroussilhe, Paul Barham, Yiding Jiang, Theodoros Rekatsinas, Gustavo Alonso, and Foteini Strati for helpful discussions.
This work was supported as part of the Swiss AI Initiative by a grant from the Swiss National Supercomputing Centre (CSCS) under project ID a09 on Alps.
Maximilian Böther is supported by the Swiss National Science Foundation (project number 200021\_204620).
\end{acks}

\bibliographystyle{ACM-Reference-Format}
\bibliography{paper}

\end{document}

%% file: meta/packages.tex
\usepackage[utf8]{inputenc}
\usepackage[T1]{fontenc}
\usepackage{mathtools}
\usepackage{balance}
\usepackage{enumitem}
\usepackage{stfloats}
\usepackage{algpseudocode}
\usepackage{amsmath}
\usepackage[frozencache,cachedir=.]{minted}

\usepackage{booktabs}
\usepackage{color, colortbl}
\usepackage[ruled, linesnumbered, vlined]{algorithm2e}
\PassOptionsToPackage{hyphens}{url}
\usepackage{adjustbox}
\usepackage{microtype}
\usepackage{listings}
\usepackage{svg}
\usepackage{float}
\usepackage{multirow}
\usepackage{subfig}
\usepackage{graphicx}
\usepackage
[
    noabbrev,   %
    nameinlink, %
]{cleveref}

\usepackage[normalem]{ulem}
\useunder{\uline}{\ul}{}

\usepackage{csquotes}
\usepackage{meta/dynamicwidth}

%% file: meta/commands.tex
\newcommand{\revision}[1]{%
  \PackageError{revision}{\string\revision\space is still used}{%
    Remove all instances of \string\revision\space (search for "\string\revision{...}").%
  }%
}

\makeatletter
\newcommand\thefontsize{The current font size is: \f@size pt}
\makeatother

\newcommand*{\mixtera}{\texorpdfstring{\textsc{Mixtera}}{Mixtera}} %

\definecolor{LightGray}{gray}{0.95}

\newcommand{\s}{\small}

\newcommand{\Crefwl}[2]{\Cref{#1}#2}
\newcommand{\ts}{\textsc}

\definecolor{RedOrange}{rgb}{1.0, 0.31, 0.0}
\definecolor{PineGreen}{rgb}{0.0, 0.47, 0.44}
\definecolor{NavyBlue}{rgb}{0.0, 0.27, 0.65}

\newcommand{\revA}[1]{#1}
\newcommand{\revB}[1]{#1}
\newcommand{\revC}[1]{#1}
\newcommand{\revM}[1]{#1}

%% file: content/1_abstract.tex
State-of-the-art large language and vision models are trained over trillions of tokens that are aggregated from a large variety of sources.
As training data collections grow, manually managing the samples becomes time-consuming, tedious, and prone to errors.
Yet recent research shows that the data mixture and the order in which samples are visited during training can significantly influence model accuracy.
We build and present \mixtera, a data plane for foundation model training that enables users to declaratively express which data samples should be used in which proportion and in which order during training.
\mixtera{} is a centralized, read-only layer that is deployed on top of existing training data collections and can be declaratively queried.
It operates independently of the filesystem structure and supports mixtures across arbitrary properties (e.g., language, source dataset) as well as dynamic adjustment of the mixture based on model feedback.
We experimentally evaluate \mixtera{} and show that our implementation does not bottleneck training and scales to 256 GH200 superchips.
We demonstrate how \mixtera{} supports recent advancements in mixing strategies by implementing the Adaptive Data Optimization (ADO) algorithm in the system and evaluating its performance impact.
We also show how \mixtera{} enables exploring the role of mixtures for vision-language models, which is a growing area of research.

%% file: content/2_intro.tex
\section{Introduction}\label{sec:intro}

Large language and vision models (LLMs/VLMs, often called foundation models) have become omnipresent in our daily lives.
They show enormous capabilities in a diverse set of tasks~\cite{Brown2020LLMFewShot,Bommasani2022FMs,OpenAI2024GPT4,dubey2024llama3herdmodels,radford2019language}, such as assistance with writing and coding,  video understanding, and even agentic interaction with the world.
The training of such language and vision models presents new challenges for managing training data due to the ever-growing sizes of models and datasets.
To achieve high accuracy, state-of-the-art models train over trillions of tokens.
For example, Meta's Llama 3.3 70B model is trained on a corpus of 15 trillion tokens~\cite{dubey2024llama3herdmodels,Llama33ModelCard}.
These tokens typically come from aggregated data collections such as RedPajama~\cite{together2023redpajama}, Dolma~\cite{Soldaini2024Dolma}, or FineWeb~\cite{Penedo2024FineWeb}, which include data from various sources, such as Wikipedia or Common Crawl dumps.

\begin{figure}
    \begin{adjustbox}{trim=0cm 0.4cm 0cm 0cm}
                \includesvg[width=\linewidth]{img/intro_hellaswag.svg}
    \end{adjustbox}
    \caption{Dynamically adjusting the mixture using the ADO algorithm improves pre-training performance on HellaSwag over the default static mixture across model scales.}
    \vspace{-0.6cm}
    \label{fig:intro-motivation}
\end{figure}

\revA{The composition of training data is critical to model quality~\cite{Chen2024DataJuicer}.}
Hence, selecting the right proportions of data with particular characteristics (e.g., language, topic, source) has become an active area of research to improve model performance without increasing training compute budget~\cite{chen2023skillit,ye2024datamixinglaws,xie2024doremi}.
For example, Hugging Face's \textsc{SmolLM2} model is trained with four stages of data mixtures that combine web text with specialized math,
code, and instruction-following data in varying proportions~\cite{Allal2025SmolLM2}.
Algorithms such as Adaptive Data Optimization (ADO)~\cite{Jiang2024ADO}, \textsc{Aioli}~\cite{Chen2024Aioli}, \textsc{PiKE}~\cite{Li2025Pike}, and \textsc{Skill-It}~\cite{chen2023skillit} even propose adjusting the data mixture \textit{dynamically} based on the model behavior (e.g., loss per domain) during training.
~\Cref{fig:intro-motivation} shows that ADO increases the accuracy of 1.6B and 3.6B Llama-models (c.f.~\Cref{tab:models}) compared to using a static mixture on the downstream HellaSwag benchmark~\cite{Zellers2019HellaSwag}.

\begin{figure}
   \centering
   \subfloat[a][Without \mixtera: More developer, CPU, and disk resources needed.]{\includegraphics[width=0.95\linewidth]{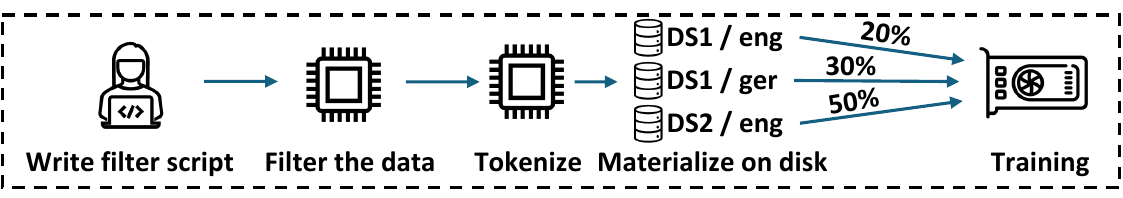} \label{fig:intro-worklfow:a}} \\
   \vspace{-0.3cm}
   \subfloat[b][With \mixtera: Declarative mixture specification and low overhead.]{\includegraphics[width=0.95\linewidth]{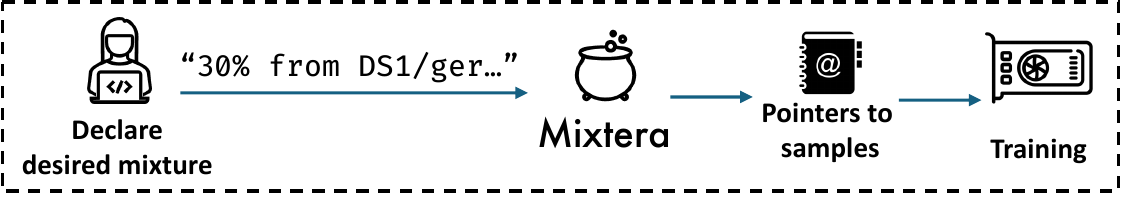} \label{fig:intro-worklfow:b}}
   \vspace{-0.1cm}
   \caption{Data preparation workflows.} \label{fig:intro-worklfow}
   \vspace{-0.65cm}
 \end{figure}

However, the process of composing training data mixtures today is manual, ad hoc, and error-prone (\Cref{fig:intro-worklfow:a}).  
Training data is typically stored on distributed filesystems in GPU clusters or data lakes in the cloud. 
ML engineers and researchers write ad hoc scripts to process the training data, filter relevant subsets with the properties of interest, often pre-tokenize it, and then mix it for their use case.
For example, a model developer may want to train on 50\,\% data from Wikipedia and 50\,\% from movie subtitles. 
This can quickly become more complex as training data may need to be mixed based on multiple characteristics.
For example, in addition to source data proportions (Wikipedia vs. movie subtitles), the developer may also want the training data to be 80\,\% in English and 20\,\% in German. 
When using mixing algorithms such as ADO, the data mixture may also need to be adjusted on the fly during training.

Model developers currently lack an open-source solution to efficiently manage and declaratively query vast amounts of training data based on the characteristics of individual data elements. 
This reflects our experience working with ML researchers as part of a large-scale initiative developing open-source LLMs and our conversations with industry teams. %
Implementing data filtering and subsequent mixing requires today's model developers to manually keep track of metadata. 
At least in the open-source world, developers typically implement this as part of the directory structure of the filesystem, e.g., developers create one subdirectory per source dataset and then sample from each directory (\Cref{fig:intro-worklfow:a}).
This approach is limited because each data sample has multiple properties that can be used to determine whether it should be used for training. 
Filesystems fundamentally do not offer the right interface for managing training data and mixing, as they do not provide declarative query interfaces or a native way to track which model was trained on what data.
Running an offline data processing job with frameworks like Apache Beam or Spark to fully materialize the mixed training set for each training run makes it difficult to quickly iterate on different mixtures during the exploration phase and leads to data duplication, increasing storage costs. 
While databases offer a declarative query interface, using a fully-fledged DBMS to track data properties would burden ML engineers with database administration, schema design, and database performance tuning.
Having a simple read-only SQLite table at each training client might seem like a straightforward solution, but it cannot dynamically adjust data mixtures and it does not interface with training frameworks like Torchtitan~\cite{Liang2024Torchtitan}.
We need a flexible data plane that lets users easily find and combine data based on any characteristic they choose, without being restricted by how files are organized.
This system should also allow users to change the data mixture on the fly and work seamlessly with  current training frameworks.

\begin{table*}[b]
\caption{Feature comparison of \mixtera{} and other open-source data loaders.}
\vspace{-0.3cm}
\small
\label{tab:intro-features}
\begin{tabular}{@{}cllll@{}}
\toprule
\textbf{}                                           & \multicolumn{1}{c}{\textbf{\mixtera}}                                       & \multicolumn{1}{c}{\textbf{HF Datasets}}                                        & \multicolumn{1}{c}{\textbf{WebDatasets}}          & \multicolumn{1}{c}{\textbf{Mosaic Streaming}}                           \\ \midrule
\multicolumn{1}{c|}{\textbf{File formats}}          & \begin{tabular}[c]{@{}l@{}}jsonl(.zst), parquet,\\ webdataset\end{tabular} & \begin{tabular}[c]{@{}l@{}}jsonl(.zst), parquet,\\ webdataset, csv\end{tabular} & webdataset                                        & \begin{tabular}[c]{@{}l@{}}Mosaic Data Shard,\\ jsonl, csv\end{tabular} \\
\multicolumn{1}{c|}{\textbf{Static filtering}}      & {\color[HTML]{2C7BB6} declarative}                                         & {\color[HTML]{FDAE61} using map UDFs}                                           & {\color[HTML]{FDAE61} using map UDFs}             & {\color[HTML]{FDAE61} using map UDFs}                                   \\
\multicolumn{1}{c|}{\textbf{Static mixtures}}       & {\color[HTML]{2C7BB6} on all properties}                                   & {\color[HTML]{FDAE61} on filesystem dirs.}                                      & {\color[HTML]{FDAE61} on filesystem dirs.}        & {\color[HTML]{FDAE61} on filesystem dirs.}                              \\
\multicolumn{1}{c|}{\textbf{Dynamic mixtures}}      & {\color[HTML]{2C7BB6} on all properties}                                   & {\color[HTML]{D7191C} no}                                                       & {\color[HTML]{D7191C} no}                         & {\color[HTML]{D7191C} no}                                               \\
\multicolumn{1}{c|}{\textbf{Native 3D parallelism}} & {\color[HTML]{2C7BB6} yes}                                                 & {\color[HTML]{FDAE61} yes, manual rank handling}                                & {\color[HTML]{FDAE61} data parallel only}         & {\color[HTML]{FDAE61} yes, for specific rank order}                     \\
\multicolumn{1}{c|}{\textbf{Checkpointing}}         & {\color[HTML]{2C7BB6} yes}                                                 & {\color[HTML]{FDAE61} using TorchData}                                          & {\color[HTML]{FDAE61} by replaying, not natively} & {\color[HTML]{2C7BB6} yes}                                              \\ \bottomrule
\end{tabular}
\end{table*}

We present \mixtera, a data plane that can be deployed on top of existing LLM/VLM training data collections stored in distributed filesystems or cloud data lakes.
\mixtera{} follows a client-server model.
\revA{The user indexes all sample metadata at the server once.
During a training job, the server statically filters out the relevant samples using SPJ-style predicates, and then continuously distributes \emph{chunks} to the clients.}
Chunks are fixed-size collections of \emph{pointers} to samples (\Crefwl{fig:intro-worklfow:b}) in files that adhere to the current mixture.
The clients fetch the relevant samples and return the relevant data to the training loop.
By storing metadata, and deferring reads to the clients, \mixtera{} scales to real-world datasets.
\revC{Chunks also enable dynamic data mixing and avoid re-materializing mixed data.}
This approach also improves the iteration speed for model developers.
While existing data loaders, as outlined in~\Cref{tab:intro-features}, rely on sampling from filesystem directories, \mixtera{} supports arbitrary properties independent of the filesystem, and makes it easy for model engineers to experiment with different filter criteria, fixed learning curricula~\cite{Allal2025SmolLM2,Xu2024Currciulum}, and fully dynamic mixing algorithms that learn the mixture during training~\cite{Jiang2024ADO,chen2023skillit,Chen2024Aioli}.
We design and implement \mixtera{} tailored to the needs of foundation model training, and contribute the following:
\begin{enumerate}[leftmargin=*,nosep]

\item \mixtera{} indexes all samples and their properties, enabling users to declaratively specify mixtures across properties independent of the filesystem structure and training frameworks.

\item \mixtera{} enables dynamically changing the data mixture and the properties used for mixing. 
It achieves this by generating and streaming chunks, i.e., fixed-size lists of pointers to samples following the current mixture.

\item \mixtera's data fetching scales to meet the ingestion throughput demands of large-scale training jobs.
We run benchmarks spanning 256 GH200 superchips and show that \mixtera{} does not limit training throughput.
As an example for how it enables model accuracy improvements, we demonstrate how to implement the ADO dynamic data mixing algorithm in \mixtera{} and its positive impact on model accuracy.

\end{enumerate}

%% file: content/3_background.tex
\section{Background}\label{sec:background}

Foundation models are large-scale deep learning models suitable for a variety of tasks~\cite{Bommasani2022FMs,CMA2013FM}.
We focus on text-generation models, i.e., autoregressive large language models (LLMs) and multimodal vision-language models (VLMs).
As of 2025, most such models are based on the Transformer architecture~\cite{Vaswani2017Attention}.
They are trained on vast corpora of training data in a self-supervised manner to maximize the likelihood of predicting the tokens of a training sequence.

\textbf{Training phases.} Training is structured into \emph{pre-training} and \emph{post-training} phases.
In pre-training, we train a randomly initialized model on a general-purpose data corpus (\Cref{subsec:back-datamix}) to derive a \emph{base model}.
In post-training, common steps include \emph{supervised finetuning} (SFT) and \emph{alignment}.
In this paper, we perform only pre-training experiments due to space constraints, but \mixtera{} can also be used for post-training.

\textbf{Distributed training.} Training foundation models requires distributing computation across multiple GPUs. %
Training frameworks typically employ 3D parallelism~\cite{BenNun2019Parallel,Hoefler20243D}, consisting of pipeline parallelism (PP), i.e., partitioning the model layers between devices~\cite{Narayanan2021Megatron,Huang2019GPipe,Narayanan2019PipeDream}, tensor parallelism (TP), i.e., splitting individual tensor operations within layers across devices~\cite{Shazeer2018MeshTF,Shoeybi2019Megatron}, and data parallelism (DP), i.e., replication of the model across device groups.
PP and TP together are referred to as \emph{model parallelism}.
Nodes within the same DP group process identical inputs, while nodes across DP groups receive different data.
As an extension to DP, fully-sharded data parallelism (FSDP) shards model parameters, gradients, and optimizer states across data-parallel workers~\cite{Zhao2023FSDP}.

\subsection{Training Data and Data Mixing}\label{subsec:back-datamix}

Pre-training data stems from data collections that include samples from various sources (e.g., Wikipedia, Common Crawl dumps, or arXiv papers).
Public examples of such collections include
RedPajama~\cite{together2023redpajama}, Dolma~\cite{Soldaini2024Dolma}, and FineWeb~\cite{Penedo2024FineWeb}.
Besides aggregating data from different sources, data engineers typically clean the data, which usually involves deduplicating, filtering (e.g., removing personal identifiable information), and applying classifiers to the data samples (e.g., to obtain a toxicity score for each sample)~\cite{Longpre2024Guide, Penedo2024FineWeb}.

\textbf{Data properties and mixtures.} 
Each data sample has properties, such as its source (e.g., Wikipedia) or its language (e.g., English).
ML engineers need to define a \emph{data mixture}, which describes how the data is combined based on its characteristics, e.g., we can train on 50\,\% data from Common Crawl and 50\,\% from movie subtitles. 
The data can be combined based on multiple characteristics simultaneously.
For instance, besides Common Crawl and movie subtitles, we might also use 80\,\% French and 20\,\% Italian data.

\textbf{Mixing algorithms.}
Selecting the best mixture is critical for model performance~\cite{chen2023skillit,ye2024datamixinglaws,xie2024doremi,Shen2024SlimPajamDC}. 
We differentiate \emph{static mixtures}, i.e., mixtures that remain constant over the entire training job, and \emph{dynamic mixtures}, i.e., mixtures that change during the training job.
Recent research proposes several algorithms for finding the best static mixture or how to adjust the mixture dynamically during training.
Algorithms such as \textsc{DoReMi}~\cite{xie2024doremi} or the data mixing laws~\cite{ye2024datamixinglaws} find a static mixture via small proxy models.

Curriculum learning is an example of a pre-defined dynamic mixture.
Xu et al.~\cite{Xu2024Currciulum} order samples from easy to hard to improve alignment.
The \textsc{SmolLM2} model was trained on 4 stages of mixtures~\cite{Allal2025SmolLM2}.
Multilingual models are often first trained on English data, followed by samples from other languages~\cite{Richburg2024MLFinetune,Xu2024MultilingualShift}.

Beyond such pre-defined schedules, for text-only models, there is also work on adapting the data mixture to the model training dynamics, e.g., by increasing the weight of data domains that have high loss.
Albalak et al.~\cite{Albalak2024OnlineMixing} model mixture components as arms of a multi-armed bandit.
\textsc{Skill-it} orders \enquote{skills} based on model feedback~\cite{chen2023skillit}.
\textsc{Aioli} builds upon \textsc{Skill-it} and provides a unified framework for estimating the best mixture during training~\cite{Chen2024Aioli}.
\textsc{PiKE} relies on gradient interactions~\cite{Li2025Pike}.
In this paper, we use Adaptive Data Optimization (ADO)~\cite{Jiang2024ADO} as an example of a dynamic mixing algorithm.

\subsubsection{Adaptive Data Optimization} \label{subsubsec:back-ado}

Adaptive Data Optimization (ADO) is a dynamic mixing algorithm that adjusts the data mixture during training based on the model's learning progress on each domain~\cite{Jiang2024ADO}. 
The key idea is to prioritize domains where the model shows rapid improvement while considering how much each domain benefits from its own samples.
ADO uses neural scaling laws to model how the loss $L_k$ of each domain $k$ decreases with the number of training samples $n$. 
To this end, it fits a power law $\hat{L}_k(n) = \varepsilon_k + \beta_k n^{-\alpha_k}$ \emph{for each domain}.
Here, $\varepsilon_k$ represents the irreducible loss of the domain, $\beta_k$ is a scaling factor, and $\alpha_k$ determines how quickly the loss decreases.
The parameters are re-fitted during training.
The algorithm combines two components to determine the mixture weights.
First, it estimates the learning speed for each domain using the derivative of the scaling law.
Second, ADO maintains a credit assignment score $\lambda_k(t)$ that indicates how much each domain contributes to its own progress, based on its recent sampling frequency.
These components are combined with a prior (initial) distribution $\mu_k$ to compute an intermediate preference distribution  $\rho_k(t)$.
To ensure stability, the final distribution $\pi_k(t)$ is then computed as a weighted average between $\rho_k(t)$ and $\pi_k(t)$'s temporal average.
Additionally, ADO enforces a minimum sampling probability for each domain.

%% file: content/4_challenges.tex
\section{Current Challenges}\label{sec:status_quo}

We identify three challenges in the status quo of training data management with current open-source infrastructure.

\textbf{\underline{Challenge 1:} Today's training data storage systems lack expressive, declarative APIs for data mixing.} 
Training data is typically stored and managed as files on distributed filesystems or objects on cloud storage.
As these systems are not natively built for foundation model training data~\cite{Wang2023LLMDataSurvey}, they lack operators for data selection, mixing, and ingestion into a training framework.
This also complicates lineage tracking, as there is no native way of tracking which model was trained on which data when the data is accessed via general-purpose filesystem calls.

\textbf{\underline{Challenge 2:} Preparing and materializing data mixtures with offline preprocessing limits flexibility and increases storage costs.} 
The current approach to preparing training data involves numerous manual offline steps with general-purpose data processing and scripting frameworks (\Crefwl{fig:intro-worklfow:a}).
For the offline cleaning step (\Cref{subsec:back-datamix}), ML engineers typically leverage data processing frameworks like Spark~\cite{Zaharia2016Spark}, Beam~\cite{Akidau2015DataflowBeam}, Data-Juicer~\cite{Chen2024DataJuicer}, or datatrove~\cite{Penedo2024DataTrove}.
Subsequent data mixing can happen offline or online.
Current online data loaders sample data from directories that reflect the mixing property, e.g., one directory per source dataset (\Cref{tab:intro-features}), and therefore do not support switching the property we mix on nor specifying hierarchical mixtures across arbitrary properties.
\revC{Due to the limited functionality of current online solutions, engineers often write ad hoc offline mixing scripts  that create a new mixed copy of the cleaned dataset for each training run, which can drastically increase storage usage and cost.}
As some frameworks like Megatron~\cite{Shoeybi2019Megatron,Narayanan2021Megatron} even require pre-tokenized data, this leads to duplication of the even bigger tokenized data files.
Offline mixing also drastically increases the iteration time and makes it difficult to get a quick sense of how a mixture will impact model training when exploring different mixing policies.

\textbf{\underline{Challenge 3:} Lack of support for dynamic mixtures during training.} 
How to train the best model on a given dataset is an active research area, with dynamic mixing emerging as a prominent new technique.
Offline preparation of the mixture or other approaches such as online mixing based on fixed directory weights or using a vanilla DBMS without additional infrastructure does not support dynamic mixture at all.
Additionally, even for researchers who are familiar with the latest mixing techniques, implementing modern mixing algorithms in a training pipeline is a painful, tedious, and error-prone task, as the codebases for mixing algorithms are often tailored directly to the training framework, as well as the data collection and properties used in the respective papers.
This hinders the adoption of dynamic mixing algorithms and makes researching, reproducing, and comparing difficult.

%% file: content/5_design.tex
\section{\mixtera's Design}\label{sec:design}

We address these challenges for training data management by building \mixtera, a foundation model training data plane.
We derive the following design goals for such a system.

\textbf{\underline{Goal G1:}} The system should implement a \emph{centralized} data layer that users can conveniently and declaratively query to mix data across arbitrary properties, independent of the filesystem structure.

\textbf{\underline{Goal G2:}} The system needs to be \emph{lightweight}, i.e., easily integrate into existing training deployments without requiring setting up many components.

\textbf{\underline{Goal G3:}} The system needs to ensure \emph{high-throughput}, \emph{determinism}, and \emph{reproducibility}, while being user-friendly and flexible.

\textbf{\underline{Goal G4:}} The system must support \emph{adjusting the mixture dynamically} during training.

\subsection{Data Model}\label{subsec:design-datamodel}
\revB{\mixtera{} operates on training datasets of the following structure:}

\revB{\textbf{Sample.} A sample \(s\) is the atomic unit of training data, \emph{as supplied by the user}.
In LLM training, the \emph{sample unit} is typically a text string or a pre-tokenized sequence; VLM training uses multimodal samples (e.g., text-image pairs).
Sample granularity varies in practice, ranging from short texts to full documents.
}

\revB{\textbf{Property.} A property $p$ is a named attribute of a sample, with a domain $\mathcal{D}_p$ of possible values.
Properties characterize samples and can be \textbf{single-valued} (each sample has exactly one value, e.g., $p_{\text{lang}} \in \{\text{English, French, German}\}$), or \textbf{multi-valued} (each sample may have multiple values, e.g., $p_{\text{topic}} \subseteq \{\text{science, politics}\}$).}

\revB{\textbf{Data Collection.} A  data collection $\mathcal{C}$ is a tuple $(\mathcal{S}, \mathbb{P}, \phi)$ where $\mathcal{S} = \{s_1, s_2, \ldots, s_n\}$ is a finite set of samples, $\mathbb{P} = \{p_1, p_2, \ldots, p_m\}$ is a set of properties, and $\phi: \mathcal{S} \times \mathbb{P} \rightarrow \mathcal{P}\left({\mathcal{D}_p}\right)$ maps each (sample, property) pair to a subset of values from that property's domain.
We also refer to it as the sample metadata.}

\revB{\mixtera{} defines mixtures fully independent of the filesystem--it is fully based on (logical) properties.}

\revB{\textbf{Mixture Key.} A mixture key $k$ is a partial mapping from a subset of properties $\mathbb{P}' \subseteq \mathbb{P}$ to sets of values, i.e., $k: \mathbb{P}' \rightarrow \bigcup_{p \in \mathbb{P}'} \mathcal{P}\left({\mathcal{D}_p}\right)$, with $\forall p \in \mathbb{P}': k(p) \subseteq \mathcal{D}_p$.
For mixture keys $k_1$ and $k_2$, we say $k_1$ \textbf{matches} $k_2$, if f.a. $p \in \text{dom}(k_1)$, $p \in \text{dom}(k_2) \land k_1(p) \cap k_2(p) \neq \emptyset$.
This relation is non-symmetric.}

\revB{\textbf{Component Key.} A sample $s \in \mathcal{S}$ induces a component key $\hat{k}_s$, s.t. $\forall p \in \mathbb{P}: \hat{k}_s(p) = \phi(s, p)$. }

\revB{\textbf{Data Mixture.} A data mixture $M$ is a mapping from mixture keys to target proportions:
$M: \mathcal{K} \rightarrow [0,1]$ with $\sum_{k \in \mathcal{K}} M(k) = 1$.}

\subsection{System Overview}\label{subsec:design-overview}

\begin{figure}
    \centering
    \begin{adjustbox}{trim=0cm 0.3cm 0cm 0cm}
                \includegraphics[width=\linewidth]{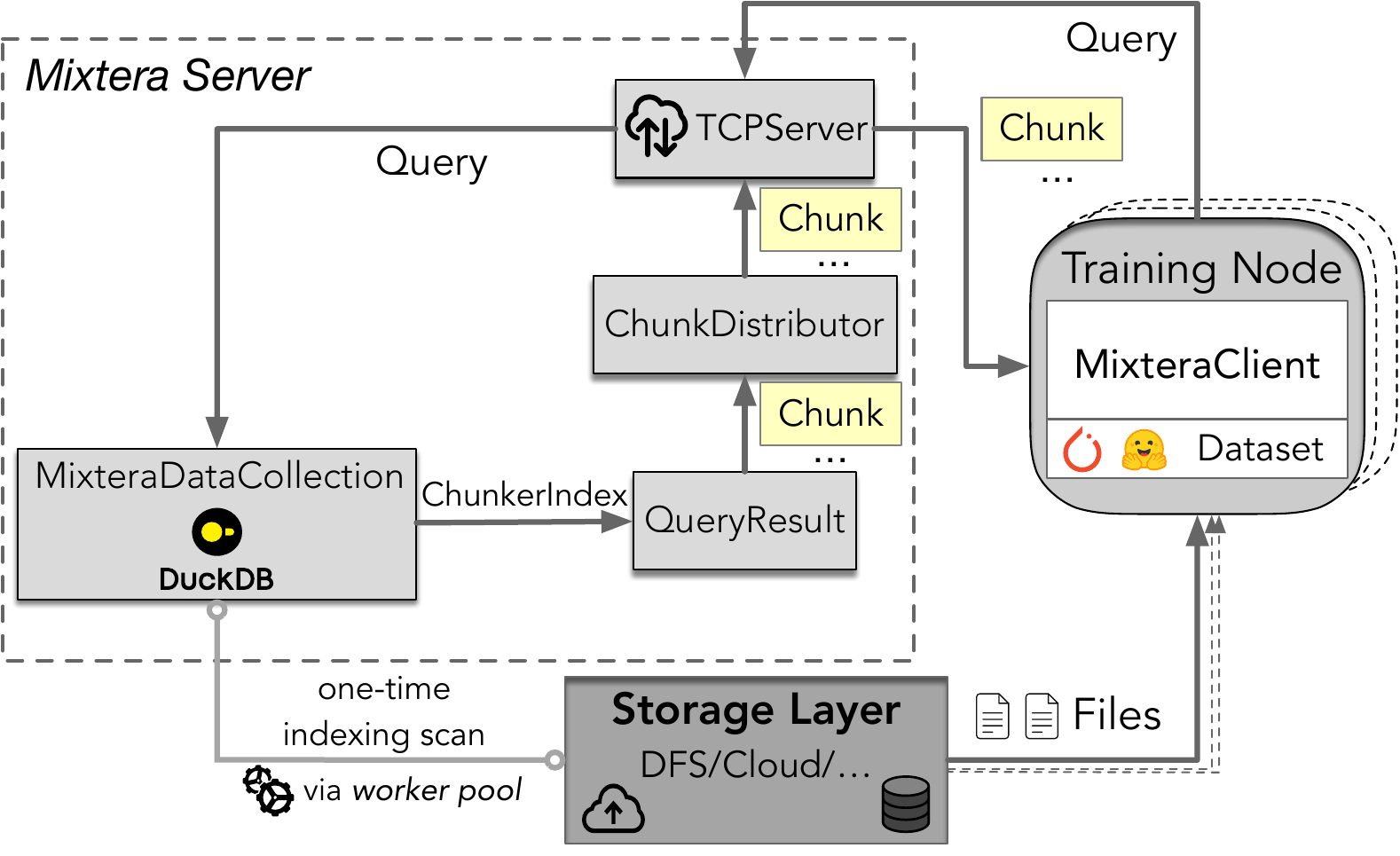}
    \end{adjustbox}
    \caption{\revA{\mixtera{} system architecture.}} 
    \label{fig:mixtera-system}
    \vspace{-0.75cm}
     \Description{TODO!}
\end{figure}

\revB{\mixtera{} is a read-only layer that can be deployed on top of existing sets of samples \(\mathcal{S}\), which are typically stored in a distributed filesystem or cloud object store.}
\Cref{fig:mixtera-system} shows the client-server architecture of the system. 
The server runs on one node, and each training node runs client instances.
\revB{\mixtera{} manages a centralized database  which stores the mapping \(\phi\), i.e., the sample metadata (G1).}
\revA{This database needs to be populated by scanning all samples once before it can be used for training. %
Since the metadata typically consists of integers and floats, it is much smaller than the actual samples.
\mixtera{} defers the reading of actual sample payloads to clients.
This enables  to scale to large-scale training datasets.
}

\revB{\mixtera{} assigns every sample a unique ID.}
It allows model developers to declaratively query the relevant samples for a training job.
To remain lightweight (G2), \mixtera{} does not reorganize or modify the data files on disk. 
It provides a standard iterator for model training that can be used in conjunction with loaders like \texttt{torch.DataLoader}. 
\mixtera{} is agnostic to the training framework, supports training interruptions using checkpoints, and ensures determinism (G3) through careful shuffling, i.e., for identical queries, \mixtera{} always provides data in an identical order, which is important for reproducibility and debugging issues like loss spikes~\cite{Zhuang2022Randomness,Cooper2022Determinism,Qian2021Seeds,Karamcheti2021Mistral,TensorflowDeterminism}.
It supports adjusting the mixture during training (G4) by transferring chunks (lists of pointers to samples) whose mixture can change over time.

\revB{\textbf{Data types.} 
\mixtera{}  supports diverse sample types (e.g., pre-tokenized text in binary files, strings in \texttt{jsonl}/ \texttt{parquet}, text–image pairs in \texttt{webdatasets}).
This covers many relevant use cases.
Even modern foundation models for tabular data treat tables as strings~\cite{Bodensohn2025Challenges}.
The sample granularity is inherent to the ingested dataset.
As \mixtera{} is a read‑only layer, it adopts whatever granularity the data already uses.
While some features such as tokenization or image decoding are modality-specific, the system is fundamentally agnostic to the sample unit.
}

\revB{\textbf{Chunks.} Chunks are \mixtera's core abstraction for scheduling and mixture enforcement. 
\emph{Abstractly}, a chunk $\mathcal{C}_h$ is a collection of samples $\left\{s_1, \ldots, s_c\right\}$ of fixed size $ c = \left|\mathcal{C}_h\right| $, and the set conforms to the current mixture $M$. That is, for each mixture key $k \in \mathcal{K}$, the proportion of samples in $\mathcal{C}_h$ matching $k$ is approximately $M(k)$.
\emph{In implementation}, \mixtera{} does not materialize or transfer the sample payloads $\mathcal{S}$.
Instead, chunks contain only \emph{pointers} to samples: metadata specifying which samples in which files to load (e.g., \enquote{samples 100-150 in \texttt{wikipedia.jsonl.zst}}).
}

\revB{\textbf{Deferred reading.} Storing only metadata and distributing pointers to samples (as opposed to ingesting the actual sample payloads into the system) has several advantages.}
First, users can store data in their locations of choice (e.g., an object store, or a distributed filesystem).
Chunks are independent of the filesystem structure (G1).
Second, it allows \mixtera{} to support dynamic mixtures (G4), as the data composition of chunks can change over time.
Third, the pointer-model avoids creating a data fetching bottleneck at the \mixtera{} server.
The server only creates chunks and each client fetches the data they need.
Fourth, we avoid a lock-in effect and allow for easy adoption on existing data collections (G2).
Last, we natively support other modalities (whereas image or video payloads would not be straightforward to ingest at scale into a database). 

\begin{figure}
\begin{minted}
[
frame=lines,
framesep=2mm,
baselinestretch=0.8,
bgcolor=LightGray,
fontsize=\footnotesize,
linenos
]
{python}
client = MixteraClient("127.0.0.1", 8080)
job_id = "test_job"
query = Query.for_job(job_id).select(("license","==","CC"))
mixture = StaticMixture(
    { MixtureKey({"language": ["JavaScript"]}): 0.7,
     MixtureKey({"language": ["HTML"]}): 0.3 },
    chunk_size=1024)
qea = QueryExecutionArgs(mixture=mixture, num_workers=4,
                         dp_groups=1, nodes_per_group=1)
rsa = ResultStreamingArgs(node_id=0, dp_group_id=0, job_id=job_id)
ds = MixteraTorchDataset(client, query, qea, rsa)
dl = torch.utils.data.DataLoader(ds, batch_size=1024, num_workers=4)

for batch in dl:
    print(batch)
\end{minted}
\vspace{-0.5cm}
\caption{An example query using \mixtera.}
\Description{TODO!}
\label{listing:mixtera-query}
\vspace{-0.55cm}
\end{figure}

\textbf{Query interface.} In~\Cref{listing:mixtera-query}, we show an \revM{example query} that statically selects only Creative Commons data, and then mixes HTML and JavaScript data in a 70:30 ratio during training.
\mixtera{} takes care of executing the query and obtaining the samples without needing to worry about correctness, even in distributed training.
The user only needs to provide the ID of the node and its data parallel group, which is obtained from the training framework.

\revA{\mixtera{} allows expressing static filter operations on properties via SPJ-style predicates}, and static as well as dynamic mixtures across all properties (G4).
\revM{The static filter defines the ground set of all potential data (e.g., only CC), while the mixture describes the proportions of how  the data from this set is mixed during training (JavaScript:HTML ratio).}
\revB{The \texttt{MixtureKey} class operationalizes its formal definition (\Cref{subsec:design-datamodel}) in a user-friendly Python API. }

\textbf{Executing a query.} %
Before submitting a query, users catalog all samples at the server. 
\revA{This involves defining the data schema as a Python class, which allows \mixtera{} workers to scan all samples and populate the database.}
Users then can submit queries.
A query is executed at the server in two phases.
First, \mixtera{} applies static filters from the query (e.g., English-only) to obtain all samples eligible for training (\texttt{QueryResult}).
Second, during training, the server distributes \emph{chunks} of that query result to the client(s), which specify which samples to train on.
The server ensures that the chunks are distributed correctly, i.e., tensor- and pipeline parallel stages receive the same input data.
The server generates chunks according to the current mixture, i.e., it iteratively takes samples from the query result such the chunk abides by the current mixture.
As an iterable data loader, \mixtera{} faces the challenges of determinism and checkpointing.
We address this by shuffling based on the query and support to load/store the query state.

\textbf{High-throughput data fetching.} The challenge with not re-organizing the user's  data files is that \mixtera{} needs to handle suboptimal data layout.
Files may have arbitrary distributions of data properties  (e.g., even if a file only contains Wikipedia data, data in different languages might be distributed randomly across files).
Formats like \texttt{jsonl} were not built with random access in mind, yet chunks force clients to load individual samples from files.
To avoid data stalls (G3), \mixtera{} reads subsequent samples in files if the samples follow the same properties (e.g., same language).
\revB{This is achieved through intervals $(f, [i, j), \hat{k})$ identifying consecutive samples in file $f$ that share the same component key $\hat{k}$.
A chunk is thus represented as a collection of intervals whose total size equals $c$ and whose union satisfies the mixture.}
We use nested CTEs in DuckDB to find these intervals as fast as possible (\Cref{subsec:design-clientside}).
Users are not required to use sequential file formats. %

\textbf{Alternative design considerations.} A metadata SQLite database at each client might seem like a simple alternative. 
\revA{While DBs can federate and query external sources with SPJ-style predicates, }
we still need an interface between the storage, query engine, and training framework, which can turn query results into input tensors (G1, G2).
Furthermore, using a database at each client would not enable dynamic mixing algorithms (G4).
\mixtera{} serves as the \emph{interface} that connects a metadata database to the training framework, with a full feature set for dynamic model training.
It also alleviates ML engineers from manually maintaining the database.
We do not find centralized metadata management to be a bottleneck even for large-scale training jobs, because the work at the service is minimal (occasional chunk creation and distribution).
\mixtera{} maintains high training throughput for large training jobs (\Cref{subsec:eval-perf}).
Industry frameworks like Google's Pathways~\cite{Barham2022Pathways} also rely on a single Python process to coordinate large-scale training.
A current limitation of \mixtera's design is that it does not enable incorporating new data into an existing query.
However, this (i) affects nearly all dataloaders as they all build some internal state, and (ii) and can in all cases be dealt with by executing a new query, e.g., in continual pretraining from a model checkpoint.

\textbf{Open source ecosystem.} 
\mixtera{} comes as a Python package that provides the entry point for the server and abstractions for the client.
The codebase, consisting of approximately 11\,k lines of Python and C++ (excluding tests), is open-source\footnote{Available at \url{https://github.com/eth-easl/mixtera}.}.
It is rigorously tested with a full set of unit and integration tests. 
We are continuing to add features and welcome contributions.

%% file: content/6_impl.tex
\section{Implementation}\label{sec:implementation}

We explain how \mixtera{} ingests sample metadata (\Cref{subsec:design-dc-mgmt}), how it executes queries and creates chunks (\Cref{subsec:design-queryexec}), how those chunks are parsed at the client (\Cref{subsec:design-clientside}), and describe \mixtera's integration into training frameworks (\Cref{subsec:design-integration}).

\subsection{Metadata Insertion}\label{subsec:design-dc-mgmt}

\revA{\mixtera{} manages the metadata $\phi$
in an \texttt{MixteraDataCollection} (MDC) which uses DuckDB~\cite{Raasveldt2019DuckDB} as the 
underlying DBMS to efficiently store and query sample properties.
The implementation itself is agnostic to the DBMS.}
\revA{While DuckDB’s pluggable extension architecture (or similar approaches like PostgreSQL’s Foreign Data Wrappers or SQLite’s Virtual Tables) would allow to query the sample files directly, to defer reading (\Cref{subsec:design-overview}) to the training nodes, the MDC serves as a metadata index.}

 \revA{\textbf{Initial ingestion.} To populate the MDC, users need to define a \texttt{MetadataParser}.
 A \texttt{MetadataParser} operationalizes the property set  $\mathbb{P}$ and extraction of $\phi(s, p)$ for each sample $s$.
 It defines the schema specifying $\mathbb{P}$ and domains $\mathcal{D}_p$ for each property.} 
 In Python, the schema is a list of properties which have a type (e.g., string or enum), a nullable field, and a multiple field, describing whether a single sample can take multiple values for this properties (e.g., several languages).
\mixtera{} adjusts the underlying database table by mapping the Python schema to a proper database schema. 
The system comes with a set of pre-defined parsers for common datasets and enables users to define custom parsers. %

\revA{\textbf{File scanning.}  
Metadata ingestion is a one-time preprocessing step.
The MDC first accumulates all data files and prepares the database schema.
\mixtera{} then parallelizes metadata extraction using a worker pool.
Workers process files in batches, where each worker sequentially reads samples from its assigned files and applies the \texttt{MetadataParser} to extract property values (e.g., parsing JSON fields).
Since DuckDB does not support concurrent insertions from multiple processes, \mixtera{} aggregates worker results in the main process.
To optimize insertion throughput, we convert the collected metadata to columnar PyArrow in-memory tables before bulk insertion into DuckDB.
Importantly, metadata extraction is \emph{decoupled} from training. 
Users must define properties upfront, but can later add new properties by re-scanning the dataset with an updated parser without affecting existing metadata.}

\subsection{Server-Side Query Execution}\label{subsec:design-queryexec}

\revM{After registering data, users can execute queries.}
When a client sends a query, the server executes it in two phases.
The first phase is performed via the MDC and applies static filters to identify all relevant samples, and groups consecutive samples into intervals.
The second phase constructs a data structure called \texttt{ChunkerIndex} that enables efficient, mixture-aware chunk generation.

\subsubsection{SQL generation and interval detection} 
After receiving an object representation of the query (c.f.~\Cref{listing:mixtera-query}), similar to ORM frameworks like sqlalchemy, the \mixtera{} server generates a base SQL query from this object.
This query returns a table in which each row represents a sample that the user is interested in.  
\mixtera{} ensures that the generated SQL matches the MDC's table schema, e.g., whether a property can have multiple values or not.

A key challenge for \mixtera{} is efficient random access to samples within files.
File formats like jsonl or parquet are optimized for sequential reading rather than random access. 
To address this, \mixtera{} implements an interval-based approach: the server wraps the base filtering query in an outer query that identifies continuous ranges of samples sharing identical properties within the same file.
Consider the following example result of a base filtering query:

\begin{center}
\begin{footnotesize}
\begin{tabular}{@{}cccc@{}}
\toprule
Sample ID & File ID & Language & License \\
\midrule
1 & 1 & JavaScript & MIT \\
2 & 1 & JavaScript & MIT \\
3 & 1 & JavaScript & MIT \\
4 & 1 & Python & Apache \\
5 & 1 & Python & Apache \\
1 & 2 & Python & Apache \\
\bottomrule
\end{tabular}
\end{footnotesize}
\end{center}

Instead of treating these as six individual samples, \mixtera{} identifies three intervals:
\begin{itemize}
    \item Interval 1: Samples 1-3 (File 1, JavaScript, MIT)
    \item Interval 2: Samples 4-5 (File 1, Python, Apache)
    \item Interval 3: Sample 1 (File 2, Python, Apache)
\end{itemize}

Even though samples 4-5 (file 1) and 1 (file 2) share the same properties (Python, Apache), they are in different files and thus form separate intervals.
The primary key is formed by the sample and file ID.
\mixtera{} constructs a SQL query that processes the data in multiple stages:

\begin{enumerate}[leftmargin=*,nosep]
    \item First, \mixtera{} establishes a Common Table Expression (CTE) named \texttt{base\_data} that contains the filtered samples:
    \begin{minted}[frame=lines,framesep=2mm,baselinestretch=0.8,bgcolor=LightGray,fontsize=\footnotesize]{sql}
WITH base_data AS (
    -- Our generated base filtering query here, e.g.,
    SELECT * FROM samples WHERE license = 'MIT'),
    \end{minted}

    \item Next, \mixtera{} identifies breaks in the sample sequence using window functions. The \texttt{grouped\_samples} CTE calculates the difference between consecutive sample IDs within groups sharing the same properties:
    \begin{minted}[frame=lines,framesep=2mm,baselinestretch=0.8,bgcolor=LightGray,fontsize=\footnotesize]{sql}
grouped_samples AS (
    SELECT *, sample_id - LAG(sample_id, 1, sample_id)
    OVER (PARTITION BY file_id, lang, license 
            ORDER BY sample_id) AS diff
    FROM base_data),
    \end{minted}
    Here, a \texttt{diff} value of 1 indicates consecutive samples, while any other value indicates a break in the sequence.

    \item The \texttt{intervals} CTE then groups the sequences into intervals:
    \begin{minted}[frame=lines,framesep=2mm,baselinestretch=0.8,bgcolor=LightGray,fontsize=\footnotesize]{sql}
intervals AS (
    SELECT file_id, lang, license,
        SUM(CASE WHEN diff != 1 THEN 1 ELSE 0 END)
            OVER (PARTITION BY file_id, lang, license 
                  ORDER BY sample_id) AS group_id,
        MIN(sample_id) as int_strt, MAX(sample_id)+1 as int_end
    FROM grouped_samples
    GROUP BY file_id, lang, license, diff, sample_id)
    \end{minted}
    The \texttt{group\_id} is incremented when there is a break in the sequence, creating unique identifiers for each interval.

    \item Finally, \mixtera{} aggregates the results to get the final intervals:
    \vspace{-0.5cm}
    \begin{minted}[frame=lines,framesep=2mm,baselinestretch=0.8,bgcolor=LightGray,fontsize=\footnotesize]{sql}
SELECT file_id, lang, license, group_id,
  MIN(int_strt) as interval_start, MAX(int_end) as interval_end
FROM intervals
GROUP BY file_id, lang, license, group_id
ORDER BY file_id, interval_start;
    \end{minted}
\end{enumerate}

Using intervals of samples can only improve I/O if samples within files are clustered by properties and not randomly distributed.
Since \mixtera{} is read-only by design, it does not re-shuffle data.

\subsubsection{Chunk generation}\label{subsubsec:impl-chunkgen}
After obtaining the query result with all relevant intervals, the server next runs the chunk generation algorithm.
This algorithm is based on the \texttt{ChunkerIndex} data structure, which organizes sample ranges by their properties.
\revB{We re-visit the \texttt{MixtureKey} concept from an implementation perspective.}

\revB{\textbf{\texttt{MixtureKey} abstraction.} A \texttt{MixtureKey} represents a set of properties and their values.
The class implements the 
formal definition from~\Cref{subsec:design-datamodel}: a partial mapping 
$k: \mathbb{P}' \rightarrow \bigcup_{p \in \mathbb{P}'} \mathcal{P}(\mathcal{D}_p)$ for 
some $\mathbb{P}' \subseteq \mathbb{P}$. 
The matching relation enables flexible querying: 
$k_1$ matches $k_2$ if $\forall p \in \text{dom}(k_1): p \in \text{dom}(k_2) \land 
k_1(p) \cap k_2(p) \neq \emptyset$.
This matching is crucial as the resulting interval table from DuckDB contains the full cross-product of all properties---a sample might have values for language, license, size, topic, and more---while a mixture specification may consider only a subset of these properties, as we discuss in the next paragraph.}
It also allows us to define mixtures on \emph{multiple properties} with \emph{multiple values}, instead of being limited to a single property (c.f.~\Cref{sec:status_quo}).
To ensure deterministic behavior, we implement a total ordering over keys based on the number of properties, property names, and their values.
We sometimes refer to a specific \texttt{MixtureKey} as a \emph{domain}, e.g., the key for \texttt{lang:English} defines the domain of English samples.

\revB{\textbf{The \texttt{ChunkerIndex}.} Recall from~\Cref{subsec:design-datamodel} that each 
sample $s \in \mathcal{S}$ induces a component key $\hat{k}_s$ defined f.a. $p \in \mathbb{P}$ as 
$\hat{k}_s(p) = \phi(s, p)$.  The \texttt{ChunkerIndex} 
organizes intervals by these component keys, i.e.,
$\text{ChunkerIndex}: \left\{\hat{k}_s \mid s \in \mathcal{S}'\right\} \rightarrow \text{DatasetID} 
\rightarrow \text{FileID} \rightarrow \text{List}\left[[i,j)\right]$, where $\mathcal{S}'$ is the filtered subset from the query, and intervals $[i,j)$ group 
consecutive samples sharing the same $\hat{k}$.
While the index maintains the complete property information of samples, it enables efficient sample lookup:
given a mixture key $k$, we identify all component keys $\hat{k}$ where $k$ 
matches $\hat{k}$, and retrieve their associated intervals.}
Consider a simplified example with \texttt{MixtureKey}s as strings.
A fragment of the \texttt{ChunkerIndex} might look like:
\vspace{-0.2cm}
\begin{minted}[frame=lines,framesep=2mm,baselinestretch=0.8,bgcolor=LightGray,fontsize=\footnotesize]{python}
{ "language:JavaScript,HTML;license:MIT": {
    ds_1: {
      file_1: [(1,4), (10,15)],  # half (right) open ranges
      file_2: [(1,2)]
    }},
  "language:Python;license:Apache": { ds_1: { file_1: [(1,2)] }}}
\end{minted}

In this example, a query for \texttt{language:JavaScript} would match the first key despite it having the additional license property and two assigned languages.
This demonstrates how the \texttt{MixtureKey} matching allow to work with the full property/value cross-product in the index while supporting mixtures on subsets of properties.

\textbf{Building the ChunkerIndex.} The index is built in parallel in a C++ extension, processing the interval table from DuckDB provided in Apache Arrow format. 
Operating on the Arrow table in Python would be too slow due to Global Interpreter Lock (GIL) constraints.
Using \texttt{multiprocessing} to circumvent the GIL would require expensive pickling of nested dictionaries.
Our C++ implementation uses multithreading and only acquires the GIL at the end. %

Each C++ worker thread maintains a local index for a subset of the data. 
For each row (interval), each worker constructs a C++-representation of the \texttt{MixtureKey}, inserts the interval into its local index under this key, maintaining sorted order within each file's interval list.
After parallel processing, the local indices are merged, combining interval lists while preserving their sorted order.
In the end, we convert the index to Python objects, which often  is  the most expensive operation of this process.

\revB{\textbf{Chunk generation.} As defined in~\Cref{subsec:design-overview}, a chunk $\mathcal{C}_h$ is 
abstractly a collection of $c$ samples conforming to mixture $M: \mathcal{K} \rightarrow [0,1]$.}
\revB{In implementation, $\mathcal{C}_h$ contains only interval pointers $(f, [i,j), \hat{k})$  rather than materializing samples.
Given $M$ and chunk size $c$, ~\Cref{alg:chunk-generation} constructs $\mathcal{C}_h$  such that for each mixture key $k \in \mathcal{K}$, the number of samples with component keys  matching $k$ approximates $M(k) \cdot c$.}
Users can set a mixture to be \emph{strict}, requiring exact proportions, or \emph{best-effort} (continue to generate chunks even if the mixture cannot be exactly fulfilled).
\revB{The algorithm iterates through each $k \in \mathcal{K}$, 
identifies all matching component keys $\hat{k}$ in the \texttt{ChunkerIndex}, and extracts intervals until the target count is reached.}
This algorithm supports dynamic mixture, as the mixture can be changed between chunks.

\begin{algorithm}[t]
\caption{Chunk generation algorithm. Some early exits and details are omitted for readability.}
\label{alg:chunk-generation}
Initialize remaining\_counts from mixture\;
chunk $\gets \emptyset$\;
progress $\gets$ true\;
\While{$\exists \text{key}: \text{remaining\_counts[key]} > 0 \text{\textbf{ and }progress}$}{
    progress $\gets$ false\;
    \ForEach{mixture\_key in remaining\_counts}{
        \ForEach{component\_key in chunker\_index}{
            \If{mixture\_key matches component\_key}{
                Take up to remaining\_counts[mixture\_key] samples from chunker\_index[component\_key]\;
                \If{$\text{got }>0\text{ samples}$}{
                    Add samples to chunk\;
                    Update remaining\_counts\;
                    progress $\gets$ true\;
                }
            }
        }
        \If{$\text{remaining\_counts[key]} > 0$ \textbf{and} is best-effort}{
            Redistribute remaining counts to other keys\; %
        }
    }
}
\If{$\forall \text{key}: \text{remaining\_counts[key]} = 0$}{
\Return chunk\;
}
\end{algorithm}

For each key in the mixture, the algorithm keeps track of how many samples we still need to put into the chunk that is currently being generated (\texttt{remaining\_sizes}).
For each key in the mixture (line 6), it checks whether it matches a component key in the chunker index (lines 7-8).
If we find a match, we try and obtain samples (ranges) from the \texttt{ChunkerIndex} for this component key (lines 9+).
A call to obtain samples for a component key can return fewer samples than requested, e.g., if we are looking for JavaScript data and we need 5 samples, but we only have 3 JavaScript/MIT licensed samples, the according component key can only return 3 samples.
\revM{Requesting $n$ samples is implemented as requesting $m$ intervals (from potentially multiple files) such that the overall number of samples in the returned list of intervals is $\leq n$.}
The lists of intervals per file are merged into the existing sorted list of intervals.

If, after traversing all component keys, we did not find sufficient samples for a key in the mixture, in strict mode, chunk generation fails.
In best-effort mode, the algorithm redistributes any unfulfilled counts to the remaining mixture components proportionally to their original ratios.
For example, if we need 100 JavaScript samples but only find 80, the remaining 20 samples would be proportionally distributed among other components.
To avoid infinite loops, we only distribute samples to keys on which we were able to find any samples in the last iteration.
This redistribution mechanism can be enabled or disabled, allowing users to either prioritize strict mixture fidelity (which may stop training when samples are exhausted) or training as long as possible with approximated proportions.

\revB{\textbf{Implementation details.} The \textit{take samples} operation (line 9) is implemented using Python generators  that yield ranges of samples and accept the number of samples needed as input through the generator's send mechanism.
This hides the complexity of range management and  allows for efficient, stateful iteration over available ranges while maintaining  control over sample counts.
There is one generator per component key that returns  ranges containing $N$ samples based on the \texttt{ChunkerIndex}, ensuring ranges are split such that excess data is never returned.
}

\textbf{Determinism.} \mixtera's implementation of this algorithm ensures determinism  because (1) the keys are processed in a consistent order and (2) when multiple component keys match a mixture key, they are considered in a deterministic order based on a seeded shuffle of all possible keys.
This ensures that identical queries with identical mixtures always produce identical chunks, which is important for debugging and reproducibility~\cite{Zhuang2022Randomness,Cooper2022Determinism,Qian2021Seeds,Karamcheti2021Mistral,TensorflowDeterminism}.

\revB{\textbf{Sampling and randomness.} Chunks are generated randomly, but not completely i.i.d.: (i) each chunk must contain the proportions specified by the mixture $M$, and (ii) all matching component keys are considered equal.
Consider a mixture key $k = \{\text{lang: JS}\}$ that 
matches component keys $\hat{k}_1 = \{\text{lang: JS, license: MIT}\}$ 
and $\hat{k}_2 = \{\text{lang: JS, license: Apache}\}$. 
The order in which $\hat{k}_1$ and $\hat{k}_2$ are processed is determined by a seeded shuffle, ensuring determinism across runs while providing randomization across different matching keys.
Once a component key is selected, samples are drawn from its intervals.
Within each file, intervals are used for I/O efficiency, but the order in which files are processed is shuffled.}

\revB{Importantly, if the sample count for mixture key $k$ is satisfied  by samples from $\hat{k}_1$, samples from $\hat{k}_2$ may never be selected for that chunk.
This is not a sampling bias but rather reflects the mixture specification.
By omitting the \textit{license} property from $k$, the user declares all JavaScript samples equivalent regardless of license.
To ensure unbiased sampling across all relevant dimensions, users should include  all properties they care about in the mixture.
}

\textbf{Mixture types.} 
All mixture classes implemented in \mixtera{} share a common interface that converts their specifications into a mapping from \texttt{MixtureKey}s to sample counts per chunk, used by the chunk generation algorithm:
\begin{itemize}[leftmargin=*,nosep]
    \item[--] \underline{Static Mixture:} Users explicitly specify fixed proportions for different property combinations (\Cref{listing:mixtera-query}).
    This supports arbitrary properties and is not limited by, e.g., directory boundaries.
    
    \item[--] \underline{Inferring Mixture:} Automatically derives mixture proportions from the data distribution in the query result:
    This is useful when users want to maintain the natural distribution of properties.
    
    \item[--] \underline{Hierarchical Mixture:} An advanced static mixture that allows specifying nested property relationships. 
    For example, users can define that 50\,\% of the data should be legal texts, and within that, 60\,\% should be in English and 40\% in French.
    \mixtera{} automatically flattens this hierarchy into appropriate \texttt{MixtureKey}s.
    
    \item[--] \underline{Mixture Schedule:} A \enquote{meta mixture} that allows for temporal changes in mixture composition by defining a sequence of mixtures that activate at specific training steps. 
    This enables curriculum learning with predefined schedules.
    
    \item[--] \underline{Dynamic Mixture:} Allows adaptation of mixture proportions during training based on feedback (e.g., loss) from the model. 
    If an algorithm is already supported by \mixtera{} (e.g., ADO), it can be used directly.
\end{itemize}

\textbf{Chunk distribution.}  
In distributed training, it is important to guarantee that all nodes within the same data parallel group operate on the same input tensors.
\mixtera's \texttt{ChunkDistributor} wraps around the chunk generation component, and hands out chunks correctly to the training nodes, i.e., the same chunks in the same order to nodes within the same group, and different chunks to nodes in different groups for data parallelism.
To this end, the clients need to register at the server with their respective node and group identifiers.
To avoid redundant serialization overhead, the distributor caches chunks in serialized form until all nodes in a group have received them.

\subsubsection{Networking}
We implement a  TCP-based client-server protocol.
The server uses Python's asyncio framework to handle multiple concurrent client connections.
The protocol is message-based, with each message consisting of a task identifier followed by task-specific payload data.
Tasks, for example, include the execution of a query or sending out a new chunk to a client.
To handle network issues gracefully, the client implementation includes automatic reconnection with exponential backoff and configurable timeouts.
Note that only small objects such as chunks and not actual training data is transferred via \mixtera, to avoid training bottlenecks.

\subsection{Client-Side Reading}\label{subsec:design-clientside}

\revB{\mixtera's client-side abstractions provide a generator that, given a chunk from the server, yields the actual sample payloads $s \in \mathcal{S}$.}
This generator follows a two-level nested iteration pattern: an outer iteration over chunks and an inner iteration over samples within each chunk.
The outer iteration hides the complexity of network transfer and chunk generation, while the inner iteration hides the complexity of going from pointers in the chunk to actual samples.
\revC{The previous section discussed the outer step, and we now discuss this inner step, i.e., how, given a chunk, we yield sample payloads.}

\revB{\textbf{Sample granularity.}
A critical aspect of \mixtera's client-side reading is understanding the \emph{sample unit}.
As defined in~\Cref{subsec:design-datamodel}, a sample $s$ is the atomic unit of training data \emph{as supplied by the user}.
The sample granularity is inherent to the ingested dataset $\mathcal{S}$.
\mixtera{} is a read-only layer and does not modify or redefine this granularity.
In practice, sample granularity varies significantly, from books to partial sentences.
}

\revB{Varying granularities creates a challenge for mixture enforcement.
If samples come in different lengths, enforcing mixture proportions at the \emph{sample level} may not reflect the actual mixture at the \emph{token level} seen during training.
A single long sample from one domain may contribute orders of magnitude more tokens than a short sample from another domain.
To address this, \mixtera{} provides three processing modes that offer different trade-offs between mixture guarantee granularity and sample utilization.}

\textbf{Processing modes.} A chunk can be processed in three mixture processing modes with different trade-offs.
The modes influence in what order samples are yielded, i.e., in what granularity the mixture is guaranteed, and whether string samples or tokenized sequences are yielded.
All modes begin by instantiating one \emph{active iterator} per property combination.
These iterators traverse files and ranges for their respective properties in a randomized order while maintaining sequential reading within consecutive ranges for I/O efficiency.

\textbf{Overall mixture mode.} This mode processes active iterators in a randomized round-robin fashion until depletion.
This ensures the mixture ratio is maintained at the chunk level.

\textbf{Window mixture mode.} This mode guarantees the mixture on a window smaller than the chunk size.
Similarly to chunk generation, we determine how many samples per property we yield within a window.
We then go through the properties in a randomized, round robin fashion until a window has been yielded, and start again.
This mode can operate in best-effort (continues after mixture cannot be guaranteed) or strict mode (stops at the first window where the mixture cannot be maintained).
In strict mode, the number of overall samples yielded from the chunk might be smaller than the chunk size.

\revB{\textbf{Tokenized mixture mode.} This mode addresses the sample granularity issue identified above by enforcing mixture ratios at the \emph{token level} rather than the sample level.}
It wraps the active iterators with a \emph{tokenizing iterator} that takes the incoming string samples, tokenizes them, and yields tokenized samples (integer lists) with the correct sequence length.
By setting the window size equal to the chunk size we guarantee that each chunk yields at least one window of \emph{tokenized samples}.
\revB{This ensures that the mixture $M$ is respected in terms of actual training tokens, regardless of the varying lengths of the underlying samples $s \in \mathcal{S}$.}

While this mode ensures precise mixture ratios at the token level, it may result in partial utilization of longer samples.
This is an inherent issue of unbalanced datasets, and while \mixtera{} provides flexibility to handle it, the best approach is to process the datasets such that samples are (roughly) of similar size.

\textbf{File reading.} The active iterators wrapped by the processing iterators shuffle the file order but maintain sequential reading within files.
This acknowledges that formats like \texttt{jsonl} and \texttt{parquet} are optimized for sequential rather than random access, and enables us to linearly iterate through the sorted lists of ranges per file.
The complexity of reading different file formats internally is hidden by abstractions for each format.
Using the \texttt{xopen} library we support both compressed and uncompressed \texttt{jsonl}.
We optimize the reading of \texttt{parquet} files by calculating and loading only the relevant the row groups, and build upon pyarrow's \texttt{parquet}-batched-reading implementation.
WebDatasets is the only format supporting random access to samples, and we implement support using the \texttt{wids} library.
The format also gives us the option to store text-image pairs, as it can contain different modalities.
To mitigate latency from initial file operations that we observed on distributed filesystems, \mixtera{} employs a prefetching iterator that uses background threads to hide file opening latency.

\textbf{Determinism.} All random operations are seeded based on the current chunk, ensuring identical behavior across nodes.
Combined with the server-side chunk generation and distribution, this guarantees that \emph{all clients within a data parallel group yield exactly the same samples in exactly the same order}.
This property is crucial for both reproducibility across runs and correctness in distributed training.
We validated this using a suite of integration tests as well as the dataloader verification test provided by nanotron.

\textbf{Dataset abstractions.} 
Training frameworks typically require specific dataset interfaces that support \texttt{multiprocessing} with worker processes.
Besides a general-purpose interface, \mixtera{} offers a class extending torch's \texttt{IterableDataset}, and a class compatible with the Hugging Face API.
Each worker process at each node operates on its own chunk.

\subsection{Framework Integration}\label{subsec:design-integration}

\mixtera{} integrates into the training framework for checkpointing and transferring model feedback (e.g., per-domain loss).

\textbf{Checkpointing.} \mixtera's API offers a function to be called on checkpoint.
To restore from a checkpoint, we need to know (i) which chunks have been handed out to which nodes and (ii) which data loader worker processes have yielded how many samples for each node.
\mixtera{} implements (i) using the \texttt{ChunkDistributor}, which caches the query and the current state on checkpoint and can restore the in-memory state of the iterators for chunk generation based on this information.
For (ii) the \texttt{MixteraTorchDataset} uses a shared memory segment to share with the main training process the status of the data loader workers.
When we restore a checkpoint, we restore the state at the server, hand out the last chunks to each worker that they were working on, and then at the workers discard the previously yielded samples.

\textbf{Training feedback.} Dynamic mixing algorithms require a loss per property domain.
\mixtera{} offers a simple function to forward this information to the server.
Users still need to adjust the training framework.
The loss implementation needs to be adjusted s.t. it is not immediately reduced but stored per domain.
The per-domain losses along all training nodes need to be synchronized, e.g., via all-reduce, before passing it to \mixtera.

%% file: content/7_eval.tex
\section{Evaluation}\label{sec:eval}
We evaluate \mixtera{} to answer the following questions:

\begin{enumerate}[leftmargin=*,nosep]%
    \item How can we integrate dynamic mixing algorithms into \mixtera{} and what role do mixtures play for model accuracy?

    \item How does \mixtera's throughput compare to other data loaders and how well does it scale?
\end{enumerate}

\begin{table}
\caption{Model configurations.}
\vspace{-0.3cm}
\label{tab:models}
\begin{adjustbox}{max width=\linewidth}
\centering
\begin{tabular}{@{}crrrrrr@{}}
\toprule
           & \multicolumn{1}{c}{\textbf{Hid. Dim.}} & \multicolumn{1}{c}{\textbf{Interm. Dim.}} & \multicolumn{1}{c}{\textbf{KV-Hds.}} & \multicolumn{1}{c}{\textbf{Q-Hds.}} & \multicolumn{1}{c}{\textbf{Layers}} & \multicolumn{1}{c}{\textbf{RoPE-$\theta$}} \\ \midrule
{\ul 162M} & 768                                    & 2\,048                                    & 12                                   & 12                                  & 12                                  & 10\,000                                    \\
{\ul 1.6B} & 2\,048                                 & 5\,464                                    & 16                                   & 16                                  & 24                                  & 10\,000                                    \\
{\ul 3.6B} & 3\,072                                 & 8\,192                                    & 8                                    & 24                                  & 28                                  & 500\,000                                   \\ \bottomrule
\end{tabular}
\end{adjustbox}
\vspace{-0.5cm}
\end{table}

We explore the first question in a dynamic mixture case study on LLMs (\Cref{subsec:eval-ado}) and a static mixture case study for VLMs (\Cref{subsec:eval-multimodal}). We explore the second question with throughput benchmarks (\Cref{subsec:eval-perf}).

\textbf{Setup.} We run experiments  on HPE Cray Supercomputing EX254n blades, each hosting two Quad GH200 nodes.
Each node contains 4 interconnected groups of a Grace CPU with 72 cores, 128\,GB DRAM, and a H100 Hopper GPU with 96\,GB of HBM.
The nodes are connected using  a 200\,Gb/s HPE Slingshot interconnect.
The machines run Ubuntu Server 24.04 LTS with kernel 5.14.21. We build on the NVIDIA NGC 25.01 container with Python 3.12, a nightly build of PyTorch 2.7, NVIDIA driver 550.54.15, and CUDA 12.8.
We add support for \mixtera{} and other data loaders on our fork of Torchtitan (commit \texttt{ae4e402})\footnote{Available at \url{https://github.com/eth-easl/torchtitan-mixtera}.}~\cite{Liang2024Torchtitan}.
Torchtitan is part of the PyTorch ecosystem and straightforward to set up. %
We use Llama3-like models with configurations in \Cref{tab:models}.
The 162M and 1.6B models are based on Jiang et al.~\cite{Jiang2024ADO}, while 3.6B follows Meta's Llama 3.2 model.
They do not have the same parameter count as torchtitan does not tie the weight embeddings.
Our training and benchmarking data is based on The Pile~\cite{Gao2020Pile}, a frequently-used data collection used for exploring data mixing~\cite{Jiang2024ADO,xie2024doremi,Albalak2024OnlineMixing}.
We split long samples with more than 1\,500 words, with max. 20\,k samples per file.

\subsection{Dynamic Mixing using ADO}\label{subsec:eval-ado}

We demonstrate how to implement dynamic mixing algorithms in \mixtera{} and their impact on model performance, taking the ADO algorithm (\Cref{subsubsec:back-ado}) as an example.

\begin{table*}
\caption{Task performance and perplexities across models and mixtures. $\uparrow$/$\downarrow$ indicate higher/lower is better.}
\vspace{-0.3cm}
\label{tab:eval-downstream-ado}
\begin{adjustbox}{max width=0.99\linewidth}
\begin{tabular}{llccccccc||cc}
\toprule
Model & Mixture & HellaSwag~$\scriptstyle\uparrow$ & WinoGrande~$\scriptstyle\uparrow$ & ARC-E~$\scriptstyle\uparrow$ & ARC-C~$\scriptstyle\uparrow$ & Lambada (OAI)~$\scriptstyle\uparrow$ & OpenBookQA~$\scriptstyle\uparrow$ & PIQA~$\scriptstyle\uparrow$ & SlimP. Perp.~$\scriptstyle\downarrow$ & Pile Perp.~$\scriptstyle\downarrow$ \\
\midrule
\multirow{3}{*}{1.6B} & ADO & \textbf{0.411} & \textbf{0.577} & \textbf{0.577} & \textbf{0.262} & \textbf{0.603} & \textbf{0.236} & \textbf{0.707} & \textbf{25.20} & \textbf{27.79} \\
 & Default & 0.380 & 0.549 & 0.543 & 0.243 & 0.551 & 0.218 & 0.695 & 26.55 & 30.72 \\
 & Natural & 0.383 & 0.556 & 0.559 & 0.247 & 0.562 & 0.214 & 0.698 & 26.04 & 29.94 \\
\midrule
\multirow{3}{*}{3.6B} & ADO & \textbf{0.449} & \textbf{0.626} & \textbf{0.601} & \textbf{0.276} & \textbf{0.624} & \textbf{0.244} & \textbf{0.732} & 22.58 & 22.26 \\
 & Default & 0.415 & 0.597 & 0.579 & 0.255 & 0.592 & 0.210 & 0.706 & 22.09 & \textbf{22.02} \\
 & Natural & 0.419 & 0.586 & 0.593 & \textbf{0.276} & 0.598 & 0.206 & 0.723 & \textbf{21.99} & 22.53 \\
\midrule
\bottomrule
\end{tabular}
\end{adjustbox}
\end{table*}

\textbf{Training setup.}
We test the 1.6B model from Jiang et al.~\cite{Jiang2024ADO} with the \texttt{EleutherAI/GPT-NeoX-20B} tokenizer
and a 3.6B model (\Cref{tab:models}) following \texttt{Llama-3.2-3B} from Meta, including its tokenizer.
We omit the results from the 162M model for brevity. %
We use a sequence length of 2048.
For ADO, we follow the codebase and discard the first 500 steps for fitting the scaling laws, start with fitting them at step 1\,000, and then re-fit the laws every 1\,000 steps with a subsampling frequency of 10.
We also follow the codebase and \enquote{use the same step size} for all domains, i.e., instead of using the count of how often a domain has been sampled to fit the scaling laws like in the paper, we average the total sample counts evenly across all domains.
We train using non-strict token-level mixtures for 30\,000 steps, using a learning rate of 0.001 with a linear warmup for 500 steps and linear cooldown for 3\,000 steps, and the AdamW optimizer.
For 1.6B, we use 64 GPUs with a microbatch size of 32, and for 3.6B we use 128 GPUs with a microbatch size of 16, resulting in a global batch size of 2048 and total 125\,B tokens.
As mixtures, we test the default weights as in DoReMi~\cite{xie2024doremi}, the natural weights as in Jiang et al.~\cite{Jiang2024ADO}, and ADO initialized with the natural weights.

\textbf{Evaluation metrics.}
We follow Jiang et al.~\cite{Jiang2024ADO} and report both downstream task performance as well as perplexity.
For downstream tasks, we report performance on HellaSwag~\cite{Zellers2019HellaSwag}, WinoGrande~\cite{Sakaguchi2021WinoGrande}, ARC-Easy and ARC-Challenge~\cite{Clark2018ARC}, Lambada OpenAI~\cite{Paperno2016Lambada}, OpenBookQA~\cite{Mihaylov2018OpenbookQA}, and PIQA~\cite{Bisk2020PIQA}.
For perplexity, we report the average unweighted token perplexity on (i) the validation set of The Pile~\cite{Gao2020Pile}, and on (ii) SlimPajama~\cite{Cerebras2023SlimPajama} as a dataset we did not train on.
We collect all metrics using EleutherAI's lm-eval-harness~\cite{Gao2023Harness}, and use the unnormalized accuracy.

\textbf{ADO algorithm performance overview.}
\Cref{tab:eval-downstream-ado} shows the performance of all models and mixtures for the final checkpoint after learning rate cooldown.
We mark in bold the best value within a model/step group.
Generally, ADO beats the static mixtures across all downstream benchmarks.
On 1.6B, it also has the best (lowest) perplexity on both SlimPajama and The Pile, while on 3.6B, the static mixtures have slightly lower perplexity.
This shows that ADO leads to better performance for downstream tasks across scales, and demonstrates the usage of a dynamic mixing algorithm in \mixtera.

Jiang et al.~\cite{Jiang2024ADO} report that on the 1.6B model, ADO sometimes performs worse than static mixtures, which we do not confirm.
Since ADO's official repository is tightly coupled with the training framework, even after corresponding with the authors, we were not able to identify the root cause of this, partly also because their code is bound to training on specific cloud instances.
\mixtera{} decouples the training framework from the mixing algorithm, which helps developers port existing algorithms to their setups (\Cref{sec:status_quo}).

\begin{figure*}
    \centering
    \begin{adjustbox}{trim=0cm 0.4cm 0cm 0cm}
                \includesvg[width=\textwidth]{img/1b_over_time.svg}
    \end{adjustbox}
    \caption{Performance of the 1.6B model on HellaSwag, OpenBookQA, and ARC-Easy, measured every 2\,500 steps.} 
    \label{fig:bm-over-time-1b}
    \vspace{-0.3cm}
    \Description{TODO!}
\end{figure*}

\textbf{Performance over time.} To demonstrate how different tasks behave over the training, we show the performance of the 1.6B model on HellaSwag, OpenBookQA, and ARC-Easy for all training checkpoints in~\Cref{fig:bm-over-time-1b}.
Every benchmark exhibits different behavior.
For HellaSwag, ADO consistently increases its margin over the static mixtures.
For OpenBookQA, ADO performs similarly to the natural mix in the intermediate checkpoints, and benefits a lot during the learning rate cooldown.
For ARC-Easy, the mixtures perform similarly, with ADO having a small edge.
This motivates future research on data mixing using \mixtera--for example, we might be able to use intermediate evaluations instead of loss to dynamically adjust the mixture.

\begin{figure}
    \centering
    \begin{adjustbox}{trim=0cm 0.4cm 0cm 0cm}
                \includesvg[width=0.95\linewidth]{img/1b_mixture.svg}
    \end{adjustbox}
    \caption{Mixture for the 1.6B model for the 6 largest domains.} 
    \label{fig:mix-over-time-1b}
    \Description{TODO!}
    \vspace{-0.5cm}
\end{figure}

\textbf{Mixture over time.} We showcase the mixture over time for the six largest domains in~\Cref{fig:mix-over-time-1b}.
We show the mixture obtained on the 1.6B model, but unlike Jiang et al.~\cite{Jiang2024ADO}, we do not observe that other models/tokenizers lead to largely different mixtures.
In all cases, the weight of GitHub and ArXiv rapidly decreases.
Notably, the more parameters, the higher the weight of Books3, and the lower the weight of PubMed Central.
On the 3.6B model, OpenWebText2's weight  surpasses PubMed Central's weight before step 5\,000, while for the 1.6B model, they only slowly approach.

\textbf{\mixtera{} implementation.} 
We implement ADO in \mixtera{} in ca. 800 LOC.
The class only handles the core algorithm, while \mixtera{} hides the complexity of the actual mixing from the algorithm implementation.
We implement it from scratch as the original implementation of ADO is fully tied to their training framework and data setup.
The original ADO implementation  updates the current mixture $\pi$ after every step and samples the next batch based on this distribution.
This is a small discrepancy to \mixtera, which can only use a new mixture when generating a new chunk.
Each chunk may then yield several batches of data with the same mixture.
We still send the per-domain losses on every training step at the client to update ADO's internal state at the server.
Whenever a new chunk is generated, the current mixture $\pi$ from ADO is queried (\Cref{alg:chunk-generation}), and the server generates a chunk  according to that mixture.
As we find in the experiments, this slight slack does not harm performance, due to the stochastic nature of sampling.

In order to use ADO, at the training nodes, the only change needed is the implementation of a per-domain loss.
For this, the loss function (e.g., cross-entropy) needs to be called without reduction, which gives a loss \emph{per token}.
As \mixtera{} provides which token belongs to which domain, we can aggregate the losses \emph{per domain}.
We then perform an all-reduce operation across all training nodes to get the global per-domain losses and send this to the server.

During development, we switched from nanotron~\cite{HFNanotron} to torchtitan~\cite{Liang2024Torchtitan}.
Notably, as  \mixtera{} is agnostic to the training framework, no changes in \mixtera{}  were required.
This showcases the benefit of having a system like \mixtera{} that decouples the mixing from the training framework.

\textbf{Takeaways.} 
Dynamic mixtures can improve model accuracy.
ADO scales beyond the 1.6B model tested in the original paper, and beats the default weights  on all benchmarks.
Our experiments also demonstrate that a synchronous algorithm, which uses a new mixture at each training step, adapts to \mixtera's chunking system.

\subsection{Multimodal LLaVA Finetuning}\label{subsec:eval-multimodal}
To showcase \mixtera's multimodal capabilities, we evaluate the impact of static mixtures for finetuning a LLaVA-style model~\cite{liu2024improvedbaselinesvisualinstruction}.
We are not aware of prior work on the impact of data mixtures on VLMs.
The LLaVA framework trains an adapter between a pre-trained image encoder and a pre-trained LLM, and then fine-tunes the adapter and LLM  on visual instruction-following data.

\textbf{Training setup.} We base our training setup on the TinyLlaVA Factory codebase by Jia et al.~\cite{jia2024tinyllavafactorymodularizedcodebase}.
We rely on a recipe from the Factory and use \texttt{google/siglip-so400m-patch14-384}~\cite{Alabdulmohsin2023Siglip400m} as the vision encoder, \texttt{TinyLlama/TinyLlama-1.1B-Chat-v1.0}~\cite{Zhang2024TinyLlama} as the LLM, and a 2-layer MLP as the adapter~\cite{FactoryRecipe}.
We train all models on one GH200 node with 4 data parallel GPUs.
For pre-training, we use a global batch size of 512 with a learning rate of 0.001, and for finetuning use a global batch size of 128 with a learning rate of 0.00002.
We use a chunk size of 256, a cosine learning rate scheduler and the Adam optimizer.
We follow Liu et al.~\cite{liu2024improvedbaselinesvisualinstruction,LlaVARepo} and pre-train the adapter on a 558\,k subset of the LAION-CC-SBU dataset with BLIP captions.
For finetuning, we follow the TinyLLava Factory~\cite{FactoryDataPrep} \enquote{LLaVA dataset} and use 665\,k samples from six datasets (COCO~\cite{Lin2014COCO}, GQA~\cite{Hudson2019GQA}, OCR-VQA~\cite{Mishra2019OCRVQA}, TextVQA~\cite{Singh2019TextVQA}, and VisualGenome (VG)~\cite{Krishna2017VisualGenome}, and LLaVA's text-only SFT annotations~\cite{liu2024improvedbaselinesvisualinstruction}).
We pre-train the adapter once, and then vary the proportions of the finetuning datasets.
We randomly generate 256 mixtures for finetuning.
Since the number of datapoints in comparison to LLM training is small, we use a best-effort mixture and ensure we go through all samples exactly once.
All models see the same data, but in a different order.

\begin{table}
\centering
\caption{VLM scores for hand-picked mixtures.}
\vspace{-0.3cm}
\label{tab:vlm-perf}
\begin{adjustbox}{max width=\linewidth}
\begin{tabular}{l|cccccc}
\toprule
\textbf{Model} & \textbf{SQA-IMG} & \textbf{TxtVQA} & \textbf{GQA} & \textbf{MME} & \textbf{MMMU} & \textbf{POPE} \\
\midrule
\textbf{Jia et al.~\cite{jia2024tinyllavafactorymodularizedcodebase}} & 64.0 & 49.6 & 58.6 & 1256.5 & 28.3 & 86.3 \\
\textbf{Infer. Mix.} & 55.88 & 37.92 & 54.63 & 1238.76 & \textbf{29.4} & \textbf{86.6} \\
\textbf{Mix. \#252} & 63.01 & 43.87 & 56.14 & \textbf{1268.05} & \textbf{30.4} & 85.7 \\
\textbf{Mix. \#107} & 58.50 & 45.18 & 58.36 & \textbf{1283.62} & \textbf{30.1} & 85.54 \\
\textbf{Mix. \#155} & 57.14 & 42.37 & 57.14 & \textbf{1290.06} & \textbf{29.3} & \textbf{86.63} \\
\bottomrule
\end{tabular}
\end{adjustbox}
\vspace{-0.5cm}
\end{table}

\textbf{Benchmarks.}
We evaluate the models on the  GQA~\cite{Hudson2019GQA}, SQA-IMG~\cite{Lu2022ScienceQAIMG}, TextVQA~\cite{Singh2019TextVQA}, POPE~\cite{Li2023POPE}, MME~\cite{Fu2024MME}, and MMMU~\cite{Yue2024MMMU} benchmarks.
We collect all metrics using TinyLLaVA Factory.

\textbf{Results.} In~\Cref{tab:vlm-perf}, we show the results reported by Jia et al.~\cite{jia2024tinyllavafactorymodularizedcodebase}, the results we obtain using the inferring mixture (\Cref{subsubsec:impl-chunkgen}), and three mixtures out of the generated mixtures that perform well.
While the inferring mixture does not achieve their reported results, this could be either due to different data dynamics in their training, or due to a different evaluation setup we cannot reproduce as their model weights are not public.
Nevertheless, in particular on MME/MMMU/POPE, the mixtures outperform the baseline.
Mix. \#252 is weighted towards TextVQA (29.1\,\%) and OCR-VQA (19.5\,\%), with equal proportions of COCO (21.9\%) and VG (21.9\,\%); GQA (4.0\,\%) and SFT annotations  (3.6\,\%) contribute minimally.
Mix. \#107 places a large emphasis on SFT annotations (28.9\,\%) and VG (27.4\,\%), followed by COCO (25.5\,\%); GQA (5.3\,\%) and TextVQA (5.1\,\%) have lower representation, while OCR-VQA (7.8\,\%) remains a minor component.
Mix. \#155 prioritizes OCR-VQA (30.3\,\%) and TextVQA (25.7\,\%), and SFT annotations (22.7\,\%); VG (13.2\,\%) and COCO (6.9\,\%) are underrepresented, while GQA (1.2\,\%) is  least utilized.
Overall, despite seeing the same samples globally, the mixtures play a big role for model accuracy.

\subsection{Throughput Benchmarks}\label{subsec:eval-perf}

We now focus on throughput and compare \mixtera{} with other data loaders across various configurations.
\revA{The goal is avoiding data stalls, i.e., training throughput should not be reduced because the GPU is waiting for data~\cite{Mohan2021DataStalls,Kuchnik2022Plumber,Murray2021tfdata,Zhao2022MetaRM, graur2022cachew, Bother2025Modyn,Robroek2026TensorSocket,Chen2025Preproc}.}
We want to show that \mixtera{} enables on-the-fly data streaming with dynamic mixing without lowering throughput in comparison to other data loaders. 
We measure throughput in tokens per second.
Note that \emph{smaller models} and \emph{more data parallelism} increase the pressure on the data loader, while larger models reduce the pressure as the training computation takes longer.
If a data loader can sustain training small models at scale on a high-end platform like the GH200, its performance is sufficient for other scenarios as well.

We run \mixtera{} with chunk sizes of 512, 1024, and 2048.
We compare to well-known state-of-the-art data loaders, i.e., the iterable \texttt{HuggingFaceDataset} by Torchtitan (\textsc{Hf-Iter}), a mapped version (\textsc{Hf-Map}), and  the Mosaic StreamingDataset~\cite{Mosaicml2022streaming} (\textsc{Mosaic}). %
The difference between \textsc{Hf-Iter} and \textsc{Hf-Map} is that similar to \mixtera, \textsc{Hf-Iter} loads and tokenizes the data on the fly, while \textsc{Hf-Map}  preprocesses all data, including tokenization.
Note that none of these data loaders support dynamic mixing (\Cref{tab:intro-features}), they just read files from front to back.
We evaluate throughput on the 162M model since larger models only lead to lower throughput, as discussed above.
We always use FSDP since Torchtitan only enables \texttt{bfloat16} training if sharding is used.
We activate compilation, disable activation checkpointing, and use fused AdamW.
We measure throughput for 30 steps, discarding the first step.
We repeat all measurements three times, i.e., in total we have 3x30 steps.
We use the Hugging Face \texttt{EleutherAI/gpt-neox-20b} tokenizer for all data loaders.
We store all data on an SSD-backed Lustre DFS.
\begin{figure}
    \centering
    \begin{adjustbox}{trim=0cm 0.4cm 0cm 0cm}
                \includesvg[width=\linewidth]{img/singlenode_boxplot.svg}
    \end{adjustbox}
    \caption{Data loader throughput depending on the number of workers. The number in brackets indicates the chunk size.} 
    \label{fig:singlenode-tp}
    \Description{TODO!}
    \vspace{-0.5cm}
\end{figure}

\textbf{Single-node.} We train on a single GH200 node with 4 GPUs.
We use 2 data parallel replicates and shards.
In~\Cref{fig:singlenode-tp} we show the throughput for the data loaders depending on the number of data workers, i.e., data loading processes from the \texttt{torch.DataLoader}.
We test up to 16 workers, where 0 workers indicate that data is loaded in the same process as the main training loop.
The plot only show results up to 4 workers  as throughput does not increase further.
Without data workers, \mixtera{} has the highest average throughput; with one worker, the other data loaders reach their peak performance.

Overall, \mixtera{} provides similar throughput to the baselines \emph{while having a much richer feature set}, e.g., dynamic mixing.
Notably, \mixtera{} has higher throughput variance than other loaders for lower number of workers.
This is due to the random access into the files.
For every sample, it needs to open the file, seek to the correct position, and load the data, instead of bulk-transferring all the data as the other data loaders can do.
This leads to the higher variance in throughput indicated by the error bars, which overall leads to  slightly lower averages.
With a higher number of workers and larger chunk sizes, the variance decreases.
When using 4+ workers, \mixtera's throughput matches the throughput of the baselines.
Additionally, increasing the chunk size also helps, and for the 0 worker case, \mixtera{} with a chunk size of 1024 or 2048 even has a higher average throughput.

This benchmark setup is quite extreme, as we train a very small 162M model on an extremely fast GPU.
For larger, state-of-the-art model sizes, the throughput differences disappear, as computation time spent within the model forward and backward passes increases.
\mixtera's client-side overhead is minimal; the primary operations (chunk parsing and data fetching) introduce negligible overhead.

\begin{figure}
\vspace{-0.2cm}
    \centering
    \begin{adjustbox}{trim=0cm 0.4cm 0cm 0.3cm}
                \includesvg[width=0.95\linewidth]{img/scaling_boxplot.svg}
    \end{adjustbox}
    \caption{Data loader throughput when increasing the number of data parallel nodes.} 
    \label{fig:eval-scaling}
    \Description{TODO!}
\end{figure}

\textbf{Scaling out.} We investigate how the data loaders scale for larger training jobs with higher data parallelism.
We scale up to 64 GH200 nodes with a total of 256 data parallel GPUs.
We find that using a maximum number of 16 replication GPUs works best.
For 4, 8, and 16 GPUs, we use half the GPUs as replication, and shard across the rest.
~\Cref{fig:eval-scaling} shows the results with 8 data workers.
All data loaders scale linearly.
With more GPUs, the throughput variance increases a bit for all systems.
We attribute this to the random assignment of nodes in the cluster by the Slurm scheduler across the 3 repetitions.
We do not test pipeline or tensor parallelism as (i) this is not necessary for the 162M model (and a larger model would only stress the data loaders less), (ii) the data loaders besides \mixtera{} do not easily support 3D parallelism, (iii) increasing data parallelism increases the load on the system more.
This demonstrates the scalability of \mixtera's single-controller design and implementation, as well as the efficiency of its chunk generation, enabling it to effectively supply chunks to all clients even at scale.

\textbf{File formats.} All previous benchmarks use uncompressed \texttt{jsonl} data.
We test compressed \texttt{jsonl} (\texttt{jsonl.zst}), \texttt{parquet}, and the \texttt{webdatasets} format in the data loaders that support them.
Notably, only \mixtera{} supports all of these formats.
We find that the underlying file format does not impact the training throughput and therefore omit a plot.
This observation holds across different numbers of data workers, where we also observe minimal performance variation among the data loaders.

\begin{figure}
    \centering
    \begin{adjustbox}{trim=0cm 0.4cm 0cm 0cm}
                \includesvg[width=\linewidth]{img/predicate.svg}
    \end{adjustbox}
    \caption{\revC{Throughput depending on the filtering predicate with 0 and 2 workers.}} 
    \label{fig:predicate-tput}
    \Description{TODO!}
    \vspace{-0.4cm}
\end{figure}

\revC{\textbf{Predicates.} The previous benchmarks select all domains within The Pile.
To show how a filtering predicate impacts throughput, 
in~\Cref{fig:predicate-tput}, we show the throughput for chunks sizes 512/2048, and 0/2 data workers, across four representative predicates: Books3 only (large samples), CommonCrawl (CC) and ArXiv (frequent domains with small and large samples), CC only (frequent but small samples), and a set of infrequent domains (\texttt{Enron Emails}, \texttt{NIH ExPorter}, \texttt{PhilPapers}, \texttt{EuroParl}, \texttt{USPTO}).}

\revC{With 0 data workers, throughput varies with the predicate; more workers stabilize throughput. 
For chunk size 512, Only CC achieves \textasciitilde587 kTokens/s, while Books3 reaches \textasciitilde776 kTokens/s.
These differences stem from different file access patterns. 
Because The Pile is shuffled, large domains are more likely to appear in clusters, reducing repeated file scans.
Domains also differ in sample sizes: large text samples yield multiple training examples from a single read, lowering total data transfer, as seen with Books3. 
With two workers, \mixtera{} masks these effects and throughput is uniform.}

\begin{figure}
    \centering
    \begin{adjustbox}{trim=0cm 0.4cm 0cm 0cm}
                \includesvg[width=\linewidth]{img/clustering.svg}
    \end{adjustbox}
    \caption{\revC{Throughput on clustered vs. regular data.}} 
    \label{fig:clustering-tput}
    \Description{TODO!}
    \vspace{-0.4cm}
\end{figure}

\revC{\textbf{Clustering.} If data is clustered, \mixtera{} can generate longer intervals and thereby reduce the amount of data transfers.
\Cref{fig:clustering-tput} compares the throughput of the regular benchmarking data with a clustered version that sequentially writes out each domain's data.
For a small number of workers,  we generally observe an increase in avg. throughput and a slight decrease of variance. 
As before, using more workers hides these effects.
}

\subsection{\revC{Data Ingestion}}\label{subsec:eval-ingestion}
\begin{figure}
    \centering
    \begin{adjustbox}{trim=0cm 0.4cm 0cm 0cm}
                \includesvg[width=\linewidth]{img/ingestion_time.svg}
    \end{adjustbox}
    \caption{\revC{Ingestion time breakdown on 10\,\%, 100\,\%, and 250\,\% versions of The Pile with 1 and 10 attributes (3 run average).}} 
    \label{fig:ingestion-time}
    \Description{TODO!}
    \vspace{-0.6cm}
\end{figure}

\revC{Before executing queries, the user must ingest metadata (c.f.~\Cref{subsec:design-dc-mgmt}).
During ingestion, \mixtera{} reads all samples, extracts their metadata, and inserts it into the underlying metadata database.
This cost is paid \emph{only once}; thereafter, arbitrarily many queries can be executed.
In~\Cref{fig:ingestion-time} we report time spent reading the samples (scanning all files) and inserting  metadata.
We benchmark this for varying dataset sizes (10\,\%, 100\,\%, and 250\,\% of The Pile) and number of attributes (randomly generated to avoid correlation).
Other operations (e.g., parser setup) have negligible overhead.}

\revC{The timings are obtained using \mixtera's optimized default settings: multithreaded file scanning (MT) and an insertion chunk size of 2,000.
Disabling MT increases the ingestion time from 315 to around 3,000 seconds for the standard 100\,\% / 1 attribute case.
With MT enabled, decreasing the chunk size to 500 increases ingestion time from 315 s to 340 s.
Chunk sizes larger than 2,000 yield only marginal benefit and increase the risk of out-of-memory errors. 
}

\subsection{Query Execution Latency Breakdown}\label{subsubsec:eval-exec}

\begin{figure}
    \centering
    \begin{adjustbox}{trim=0cm 0.4cm 0cm 0cm}
                \includesvg[width=\linewidth]{img/execution_time.svg}
    \end{adjustbox}
    \caption{Execution time breakdown on 10\,\%, 100\,\%, and 250\,\% versions of The Pile with 1 and 10 attributes (3 run average).} 
    \label{fig:execution-time}
    \Description{TODO!}
    \vspace{-0.6cm}
\end{figure}

When a query is executed in \mixtera, the main steps are querying the \revM{populated ODC (DuckDB), building  the \texttt{ChunkerIndex}, and, optionally, writing the first checkpoint.}
In~\Cref{fig:execution-time}, we give a time breakdown for varying dataset sizes (10\,\%, 100\,\%, and 250\,\% of The Pile) and number of properties (randomly generated to avoid correlation)\revA{, following~\Cref{subsec:eval-ingestion}.}
For the 100\,\% / 1 attribute case, DuckDB takes 32\,s, and preparing the index takes 16\,s due to our C++ implementation.
Persisting the initial state checkpoint of the query dominates the runtime with 282\,s.
This is because we have to serialize nested Python dictionaries, which is slow despite engineering optimizations.
Importantly, after this initial checkpoint, future checkpoints at the server can be stored \emph{within milliseconds}, since we do not have to serialize the index again.
If no \mixtera{} checkpoints are needed, the serialization can be skipped.

 \textbf{Other systems.} The streaming \textsc{Hf-Iter} data loader can basically start streaming data immediately, the \textsc{Hf-Map} data loader, the default in nanotron, loads and tokenizes the data first, taking 2\,h 51\,min for The Pile 100\,\%.
However, to use streaming data loaders such as \textsc{Hf-Iter} or \textsc{Mosaic}, in many scenarios, users would also need to run more offline preprocessing, e.g., to perform static filtering, or reshuffling the data if we want to mix on a different property.
\mixtera{} avoids this offline preprocessing completely.

%% file: content/8_conclusion.tex
\section{Conclusion and Future Work}\label{sec:conclusion}

Understanding how data mixture recipes affect model quality is an active and critical area of ML research. We design \mixtera{} to enable researchers and model developers to easily and quickly train models with a variety of mixtures across arbitrary data properties and dynamically vary mixtures during training.
\mixtera{} is a declarative data plane for foundation model training that is training framework-agnostic. %
We demonstrate \mixtera's throughput and scalability, as well as the impact of mixtures on model quality for both LLMs and VLMs.
The system lays the foundation for implementing additional features, such as data lineage tracking for model training, as it has a global view of the data, \revA{and enables future research on compute-intensive modalities such as video~\cite{Ye2025SAND} and audio~\cite{Zhao2025Overlord}.}